\documentclass[runningheads]{llncs}
\usepackage{graphicx}
\usepackage{amsmath,amssymb} 
\usepackage{color}

\usepackage{adjustbox}

\usepackage[english]{babel}
\usepackage[utf8]{inputenc}
\usepackage{algorithmicx}
\usepackage{algorithm}
\usepackage[noend]{algpseudocode}
\usepackage[normalem]{ulem}
\usepackage[english]{babel}
\usepackage[utf8]{inputenc}
\usepackage{algorithmicx}
\usepackage{algorithm}
\usepackage[noend]{algpseudocode}
\usepackage[normalem]{ulem}
\definecolor{DarkGreen}{rgb}{0.0, 0.2, 0.0}

\newcommand{\bp}{{\bf p}}
\newcommand{\calM}{\mathcal{M}}
\newcommand{\calF}{\mathcal{F}}

\begin{document}
\pagestyle{headings}
\mainmatter

\def\ACCV20SubNumber{944}  

\title{3D Object Detection and Pose Estimation of Unseen Objects in Color Images with\\ Local Surface Embeddings} 
\titlerunning{3D Pose Estimation of Unseen Objects with Local Surface Embeddings}
%
\author{Giorgia Pitteri\inst{1}
\and
Aurélie Bugeau\inst{1}
\and
Slobodan Ilic \inst{3}
\and Vincent Lepetit \inst{2}
}
\authorrunning{G. Pitteri et al.}
%
\institute{Univ. Bordeaux, Bordeaux INP, CNRS, LaBRI, UMR5800, F-33400 Talence, France 
\email{\{giorgia.pitteri,aurelie.bugeau\}@u-bordeaux.fr}\\
\and
LIGM, IMAGINE, Ecole des Ponts, Univ Gustave Eiffel, CNRS, Marne-la-Vallee, France\\
\email{vincent.lepetit@enpc.fr}
\and
Siemens Corporate Technology, Munich, Germany\\
\email{sobodan.ilic@siemens.com}}

\maketitle

\begin{abstract}
We present an approach for detecting and estimating the 3D poses of objects in images that requires only an untextured CAD model and no training phase for new objects. Our approach combines Deep Learning and 3D geometry: It relies on an embedding of local 3D geometry to match the CAD models to the input images.  For points at the surface of objects, this embedding can be computed directly from the CAD model; for image locations, we learn to predict it from the image itself. This establishes correspondences between 3D points on the CAD model and 2D locations of the input images.  However, many of these correspondences are ambiguous as many points may have similar local geometries.  We show that we can use Mask-RCNN in a class-agnostic way to detect the new objects without retraining and thus drastically limit the number of possible correspondences.  We can then robustly estimate a 3D pose from these discriminative correspondences using a RANSAC-like algorithm.  We demonstrate the performance of this approach on the T-LESS dataset, by using a small number of objects to learn the embedding and testing it on the other objects. Our experiments show that our method is on  par or better than previous methods.
\end{abstract}

\section{Introduction}

Deep Learning~(DL) provides powerful techniques to  estimate the 6D pose of an object from color images, and impressive results have been achieved over the last years~\cite{Kehl17,Rad17,Tekin2018,Jafari2018,Xiang18,zakharov2019dpod}, including in the presence of occlusions~\cite{Oberweger2018,Peng18_PVNet,Hu20}, in the absence of texture, and for objects with symmetries~(which create pose ambiguities)~\cite{sundermeyer2020augmented,park2019pix2pose}. However, most of recent works focus on supervised approaches, which require that for each new object, these methods have to  be retrained on many different registered images of this object. Even if domain transfer  methods allow for training  such methods on synthetic images instead of real ones at least  to some  extent, such training  sessions take time,  and avoiding them is highly appealing.

Recently, a few methods were proposed for 3D pose estimation for objects that were not seen during training, only exploiting an untextured CAD model provided for the new objects.  This is an important problem in industrial contexts, but also very challenging as aligning an untextured CAD model to a color image remains very difficult, especially without pose prior.
In DeepIM~\cite{li2018deepim}, the authors propose a pose refiner able to perform such alignment given some initial pose, however it has been demonstrated on very simple synthetic images with constant lighting. \cite{Pitteri19} proposes to learn to detect corners by using training images of a small set of objects, and estimates the object pose by robustly matching the corners of the CAD model with the corners detected in the input image. However, this requires the object to have specific corners and a skilled user to select the corners on the CAD model. Very recently, \cite{Sundermeyer20} proposes a single-encoder-multi-decoder network to predict the 6D pose of even unseen objects. Their encoder can learn an interleaved encoding where general features can be shared across multiple instances. This leads to encodings that can represent object orientations from novel, untrained instances, even when they belong to untrained categories. Even if the idea is promising, to achieve competitive results, they need to use depth information and refine the pose with an ICP algorithm.

\begin{figure}[t]
  \begin{center}
\includegraphics[width=\linewidth]{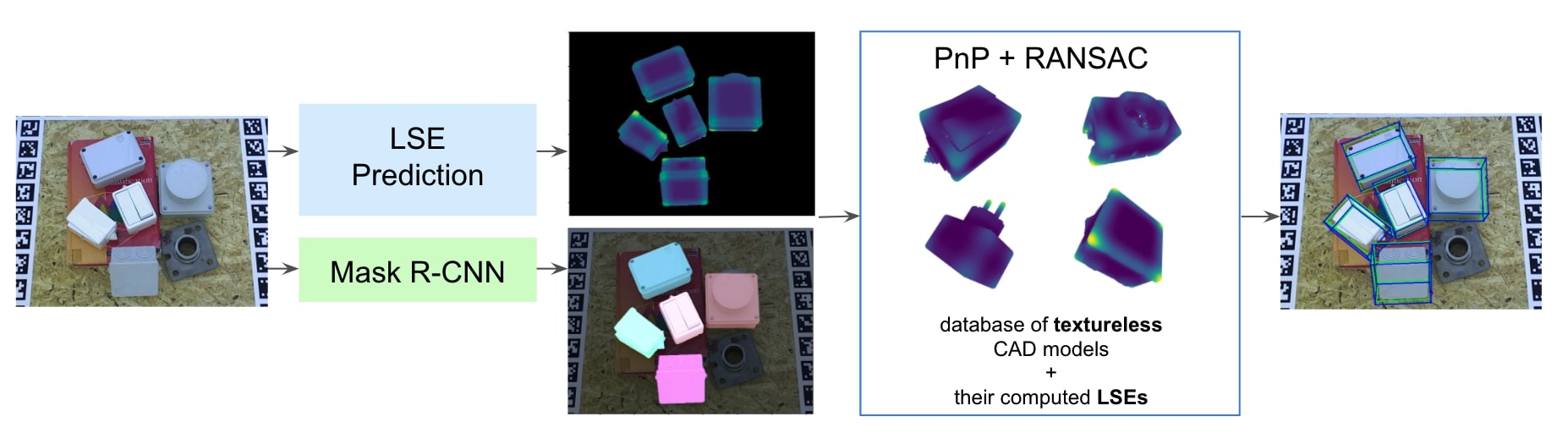}
  \end{center}
  \caption{Overview of our method. We detect and estimate the 3D poses of objects, given only an untextured CAD model, without having to retrain a deep model for these objects. Given an input RGB image, we predict local surface embeddings~(LSEs) for each pixel that we match with the LSEs of 3D points on the CAD models. We then use a P$n$P algorithm and RANSAC to estimate the 3D poses from these correspondences. We use the predicted masks to constrain the correspondences in a RANSAC sample to lie on the same object, in order to control the complexity. The LSE prediction network is trained on known objects but generalizes well to new objects. Similarly, we train Mask-RCNN on known objects, and  use mask R-CNN to segment the objects in the image. Because we train Mask-RCNN in a \emph{class-agnostic} way, it also generalizes to new objects without retraining. Note that we use masks of different colors for visualization only.}
  \label{fig:overview}
\end{figure}

In this paper, we investigate 6D object pose estimation in an industrial scenario with the challenges this implies: We want to handle symmetrical, textureless, ambiguous, and unseen objects, given only their CAD models. By contrast with some previous works, we also do not assume that the ground truth 2D bounding boxes for the objects are available. As shown in Figure~\ref{fig:overview}, our approach combines machine learning and 3D geometry: Like previous works~\cite{Brachmann16,zakharov2019dpod,park2019pix2pose,NOCS_Wang2019}, we establish dense correspondences between the image locations and 3D points on the CAD model, as they showed that this yields to accurate poses. However, there is a fundamental difference between these works and ours:  They can train a machine learning model in advance to predict the 3D coordinates of the pixels in a given image. In our case, we want to avoid any training phase for new objects. We therefore rely on a different strategy: We introduce an embedding capturing the local geometry of the 3D points lying on the object surface. Given a training set for a small number of objects, we learn to predict these embeddings per pixel for images of new objects. By matching these embeddings with the embeddings computed for 3D points on the object surface, we get 2D-3D correspondences from which we estimate the object's 6D pose using RANSAC and a P$n$P solver.

This approach is conceptually simple, robust to occlusions, and provides an accurate 6D pose. However, to be successful, some special care is needed. First, the embeddings need to be rotation invariant. Second, because of the symmetries and this rotation invariance, many correspondences between pixels and 3D points are possible \textit{a priori} and the complexity of finding a set of correct correspondences can become exponential. We control this complexity in two ways. We focus on image locations with the most discriminative embeddings as they have less potential correspondences. We also observe that Mask R-CNN~\cite{He17} is able to predict the masks of new objects when trained without any class information, and thus can segment new objects without re-training. We use this to constrain the sets of correspondences in RANSAC to lie on the same mask, and thus drastically decrease the number of samples to consider in RANSAC. 

In the remainder of the paper, we review the state-of-the-art on 3D object pose estimation from images, describe our method, and evaluate it on the T-LESS dataset \cite{Hodan17}, which is made of very challenging objects and sequences.

\section{Related Work}
 
In this  section, we first  review recent works on  3D object detection  and pose estimation from color images.  We also  review methods for using synthetic images for training as it is a popular solution for 3D pose estimation. Finally, we review the few works that consider the same problem as us.

\subsection{3D Object Detection and Pose Estimation from Color Images}

The use of Deep Learning has recently significantly improved the performance of 6D pose estimation algorithms. Different general approaches have been proposed. One approach is to first estimate 2D bounding boxes for the visible objects, and predict the 6D pose of each object directly from the image region in the bounding box~\cite{Kehl17,Rad17,Tekin2018,Xiang18,sundermeyer2020augmented}. The pose can be predicted directly using quaternions for the rotation example, or via 3D points or the reprojections of 3D points related to the object, or by learning a code book using AutoEncoders. This last method has the advantage to work well with symmetrical objects, which are common in industrial contexts. Another approach, aiming to be more robust to occlusions, is to predict for each pixel offsets to the reprojections of 3D points related to the object~\cite{Oberweger2018,hu2019segmentation,peng2019pvnet}.

Closer to our own approach, several works first predict for each pixel its 3D coordinates in the object's coordinate frame~\cite{Taylor12,Brachmann16,Jafari2018,Wang18,zakharov2019dpod,park2019pix2pose,li2019cdpn,NOCS_Wang2019,hodan2020epos}. This yields 2D-3D correspondences from which the object's 3D pose can be estimated using a P$n$P algorithm, possibly together with RANSAC for more robustness. In our case, we cannot directly predict the 3D coordinates of pixels as it can be done only for the objects or categories used for training. Instead, we learn to predict an embedding for the 3D local geometry corresponding to each pixel, and we rely on this embedding to match the pixel to its corresponding 3D point on the CAD models of new objects.

\subsection{Training on synthetic images for 6D pose estimation}

One popular approach to 6D pose estimation given only a CAD model and no, or few, real training images is to exploit synthetic images. There is however a domain gap between real and synthetic images, which has to be considered to make sure the method generalizes well to real images.

A very simple approach is to train a convolutional network for some problem such as 2D detection on real images and use the first part of the network for extracting image features~\cite{HinterstoisserPreTrainedImageFeatures2017a,Kehl17}. Then, a network taking these features as input can be trained on synthetic images. This is easy to do, but it is not clear how many layers should be used exactly. Generative  Adversarial Networks~(GANs)~\cite{Goodfellow14}  and Domain Transfer have  been used  to make synthetic images more realistic~\cite{Bousmalis16,Mueller2018,bousmalis17,zhu2017unpaired,ganin2016domain,long2015learning,tzeng2015simultaneous,lee2018diverse,Zakharov2018}.  Another interesting approach  is domain randomization~\cite{Tobin2017}, which  generates synthetic training  images with random appearance by applying drastic variations to the object textures and the rendering parameters to improve  generalization. 

These works can exploit CAD models for learning to detect new objects, however they also   require a  training phase for new objects. In this work, we do not need such phase.





\subsection{6D pose estimation without retraining}

Very few recent works already tackled 6D pose estimation without retraining for new objects. One early approach targeting  texture-less objects  is to rely on templates~\cite{Hinterstoisser12}. Deep Learning  has also  been applied  to  such problem,  by learning  to compute  a descriptor from pairs or triplets of object images~\cite{Wohlhart15,Balntas17,zakharov2017iros,bui2018icra}. Like ours, these approaches  do not  require  re-training, as  it only  requires  to compute  the descriptors for images of the new objects. However, it requires many images from points of view sampled  around the object.  It may be  possible to use synthetic images,  but then,  some  domain transfer  has  to be  performed.  But the  main drawback of  this approach is the  lack of robustness to  partial occlusions, as the descriptor is computed for whole images of objects. It is also not clear how it would  handle ambiguities, as  it is based on  metric learning on  images. In fact,  such approach  has been  demonstrated on  the LineMod,  which is  made of relatively simple objects,  and never on the T-LESS dataset,  which is much more challenging.

More recently, DeepIM proposed a pose refiner able to refine a given initial pose. In~\cite{li2018deepim}, this refiner was applied to new objects, but only on very simple synthetic images with constant lighting. \cite{Pitteri19} proposes to learn to detect corners by using training images of a small set of objects and estimates the object pose by robustly matching the corners of the CAD model with the corners detected in the input image. This method requires objects to have specific corners and to offline select corners on the CAD model. Even more recently, \cite{Sundermeyer20} proposes an extension of \cite{sundermeyer2020augmented}  able to generalize to new objects. Thanks to the single-encoder-multi-decoder architecture, they are able to learn an interleaved encoding where general features can be shared across multiple instances of novel categories. To achieve competitive results, they need to use depth information and refine the pose with an ICP algorithm. 

Our approach is related to \cite{Pitteri19}, but  considers any 3D location on the objects to get matches, not only corners. We compare against \cite{Pitteri19} and \cite{Sundermeyer20} in the experimental section.

\label{sec:related_work}

\section{Method}
\label{sec:method}

We describe our approach in this section. We first explain how we compute the local surface embeddings and how we obtain correspondences between the CAD models and the images. We then describe our pose estimation algorithm.


\subsection{Local Surface Embeddings}

To match new images with CAD models, we rely on embeddings of the local surfaces of the objects. To be able to match these embeddings under unknown poses, they need to be translation invariant and rotation invariant. Achieving translation invariance is straightforward, since we consider the local geometry centered on 3D points. Achieving rotation invariance is more subtle, especially because of ambiguities arising in practice with symmetrical objects. This is illustrated in Figure~\ref{fig:descriptors_computation}(b): We need to compute the same embeddings for local geometries that are similar up to a 3D rotation.

\newcommand{\bM}{\mathbf{M}}
\newcommand{\bP}{\mathbf{P}}
\newcommand{\bv}{\mathbf{v}}
\newcommand{\pose}{\mathbf{pose}}

\newcommand{\emb}{\text{LSE}}
\newcommand{\IoU}{\text{IoU}}
\newcommand{\proj}{\text{proj}}

More exactly, given a 3D point $\bP$ on the surface of an object, we define the local geometry as the set of 3D points $\bM_n$ in a spherical neighborhood centered on $\bP$ and of radius $r$. In practice, on T-LESS, we use $r=3cm$. To compute a rotation-invariant embedding, we transform these points from the object coordinate system to a local patch coordinate system using a rotation matrix computed from the decomposition of the covariance matrix of the 3D points $\bM_n$ after centering on $\bP$~\cite{Eggert97}:
\begin{equation}
C = \sum_n \bv_n \cdot \bv_n^\top \> ,
\end{equation}
where $\bv_n = (\bM_n - \bP)$ using a Singular Value Decomposition~(SVD):
\begin{equation}
C = L^\top \Sigma R \> .
\label{eq:LSR}
\end{equation}
%


$R$ is an orthogonal matrix, but not necessarily a rotation matrix, and small differences in the local geometry can result in very different values for $R$. We therefore apply a transformation to $R$ to obtain a new matrix $\bar{R}$ so that $\bar{R}$ is a suitable rotation matrix. It can be checked that applying $\bar{R}$ to the $\bv_i$ vectors will achieve rotation invariance for the local surface embeddings.

\begin{figure}[t]
    \begin{center}
        \begin{tabular}{cc}
            \includegraphics[width=0.4\linewidth]{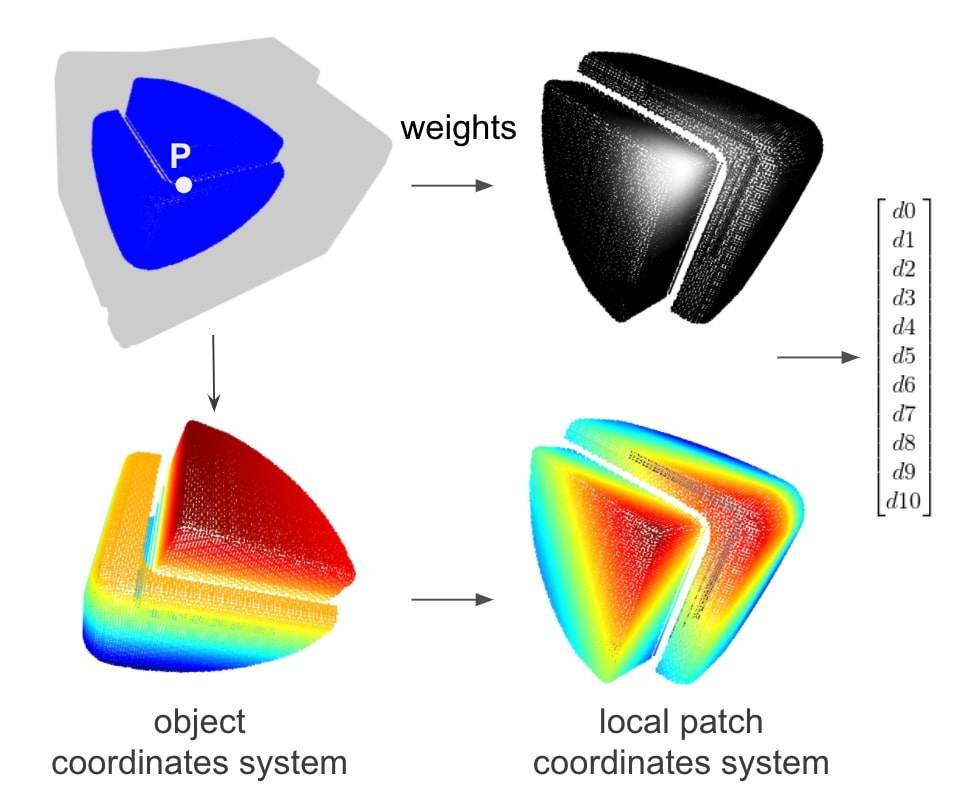} &
            \includegraphics[width=0.4\linewidth]{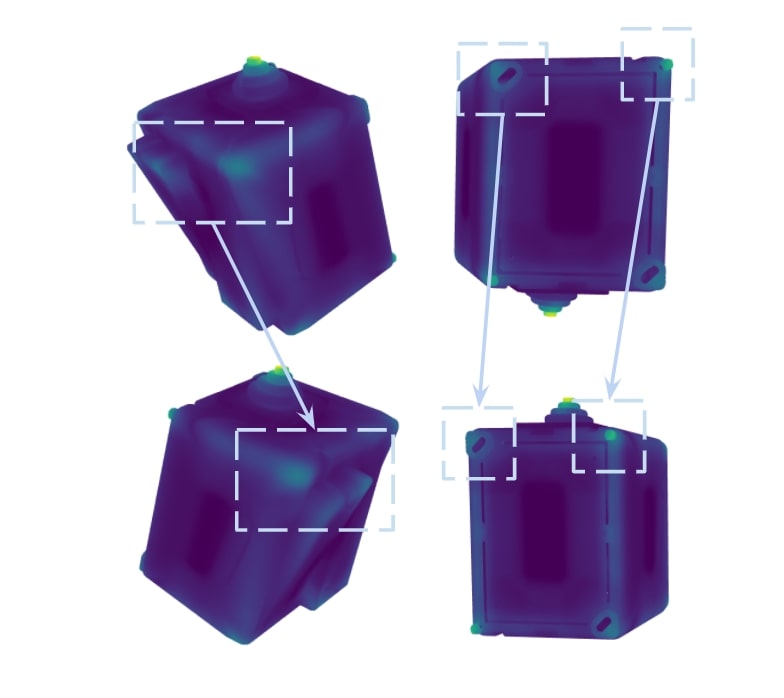} \\
            (a)  & (b)
        \end{tabular}
    \end{center}
    \caption{(a): Computation of the LSEs for a given point $\mathbf{P}$ on a CAD model. We transform the 3D points in the neighborhood of $\bP$ into a rotation-invariant local system and weight them before computing their moments. (b): Visualization of the rotation-invariance property on different parts of the same object. Similar local geometries yield similar LSEs. Through this paper, we represent the LSEs using only their 3 first values mapped to the red, green, blue channels except for Figure~\protect\ref{fig:embeddings_all_dims} that shows all the values. }
\label{fig:descriptors_computation}
\end{figure}


Let's denote by $r_1$, $r_2$, and $r_3$ the rows of $R$, and by  $\bar{r}_1$, $\bar{r}_2$, and $\bar{r}_3$ the rows of $\bar{R}$. Applying $R$ to the normal $n$ of the object's surface at $\bP$ yields a 3-vector $R \cdot n$ close to either $[0,0,1]^\top$ or $[0,0,-1]^\top$, depending on the orientation of $R$ selected for the SVD. For normalisation, we choose that $\bar{R} \cdot n$ should always be closer to $[0,0,1]^\top$. We therefore compute $o = r_3^\top \cdot n$. If $o$ is positive, we take $\bar{r}_3 = r_3$, otherwise we take $\bar{r}_3 = -r_3$.  As a result, $\bar{R} \cdot n$ is always closer to $[0,0,1]^\top$ that to $[0,0,-1]^\top$. Finally, we take $\bar{r}_1 = r_1$ and $\bar{r}_2 = -\bar{r}_1 \wedge \bar{r}_3$, where $\wedge$ denotes the cross-product, which ensures that $\bar{R}$ is a rotation matrix. 

We explain now how we define the local surface embeddings. For our experiments, we use the local moments of the local 3D points for simplicity but any other embeddings such as \cite{Deng18} could also work. Let us denote by $[x_n, y_n, z_n]$ the vectors $\bar{R} \bv_n$, then local surface embeddings can be computed as:
\begin{equation}
\emb_{i,j,k}(\bP) = \sum_n w_n x_n^i y_n^j z_n^k \> ,
\label{eq:embedding_formula}
\end{equation}
where $w_n=\exp(-\|\bv_n\|^2/\sigma^2)$ is a weight associated to each point based on its distance from $\mathbf{P}$~(we use $\sigma=5$ in practice) and $i, j, k$ are exponents in the range $[0, 1, 2]$. Theoretically it is possible to take all the combinations of exponents but we empirically found that the most discriminative values are computed using: $i\in\{0,2\}$, $j\in\{0,2\}$, $k\in\{0,1,2\}$, which gives 11 values for the full vector $\emb(\bP)$ as taking $i=j=k=0$ gives a constant value and is not useful. Finally, we normalize the values of $\emb(\bP)$ to zero mean and unit variance so they have similar ranges. Figure~\ref{fig:embeddings_all_dims} displays the embeddings for an example image.



\subsection{Predicting the local surface embeddings for new images}
\label{sec:network}

Given a new CAD model, it is trivial to compute the local surface embeddings on points on its surface. Given a new input image, we would like to also compute the embeddings for the object points visible in this image. We use a deep network to perform this task.
To do so, we create a training set by generating many synthetic images of known objects under various poses. We also compute the LSEs for all the pixels corresponding to a 3D point of one of the objects. We then train a U-Net-like architecture~\cite{Ronneberger15} to predict the LSEs given a color image. More details on the architecture and its training are provided in the experimental section.

This training is done once, on known objects, but because the embeddings depend only on the local geometry, the network generalizes well to new objects, as shown in Figure~\ref{fig:embeddings_prediction}. 

\newcommand{\w}{0.25\linewidth}
\begin{figure*}[t]
  \begin{center}
    \begin{tabular}{cccc}
      \includegraphics[width=\w]{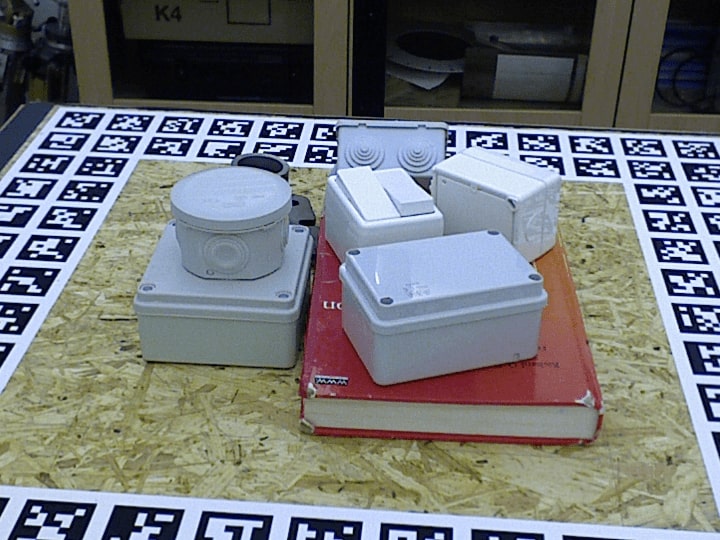} &
      \includegraphics[width=\w]{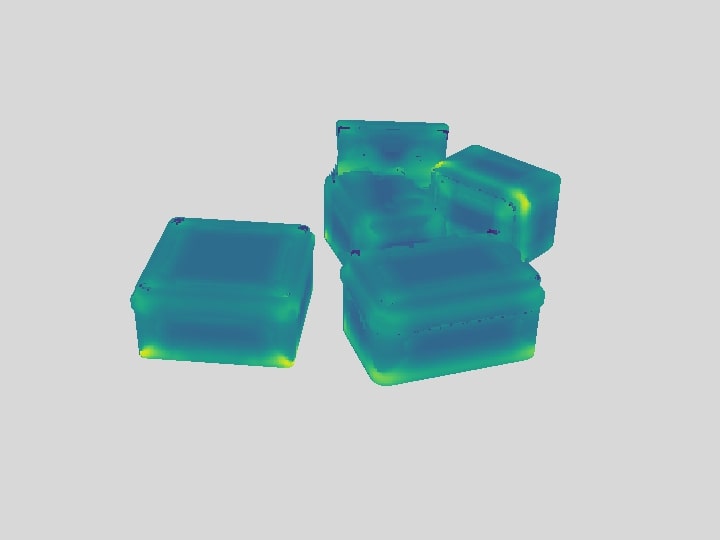} &
      \includegraphics[width=\w]{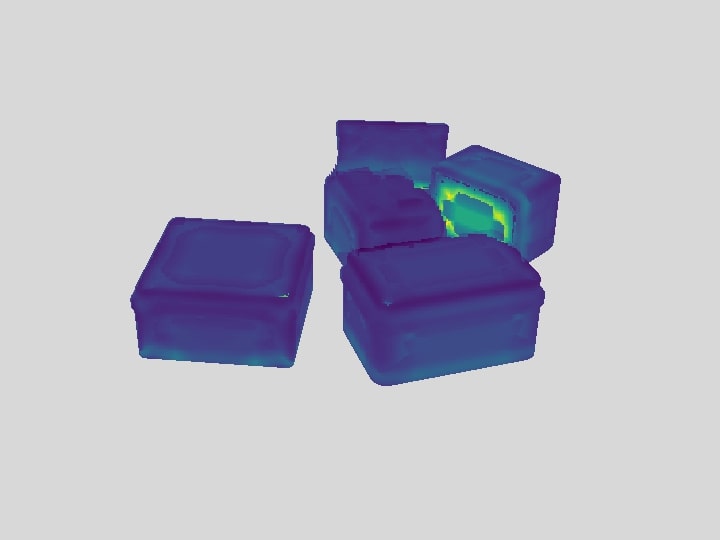} & 
     \includegraphics[width=\w]{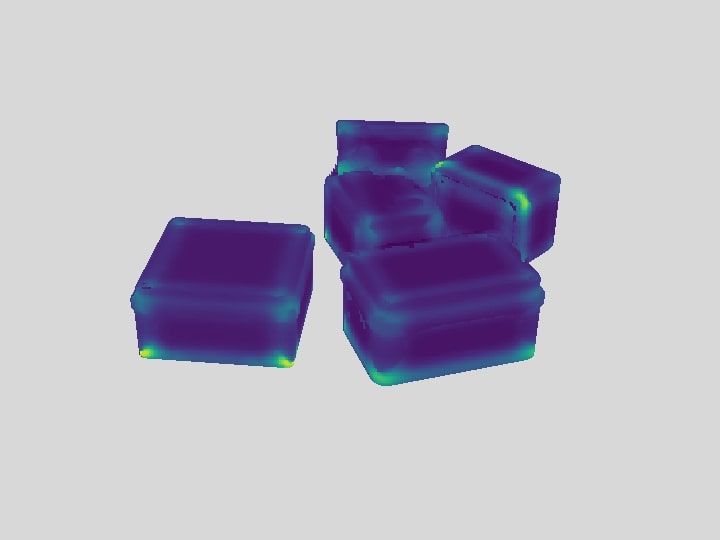} \\
       Input image & $i=0$, $j=2$, $k=1$ & $i=0$, $j=2$ , $k=0$ & $i=0$, $j=2$, $k=2$\\
      \includegraphics[width=\w]{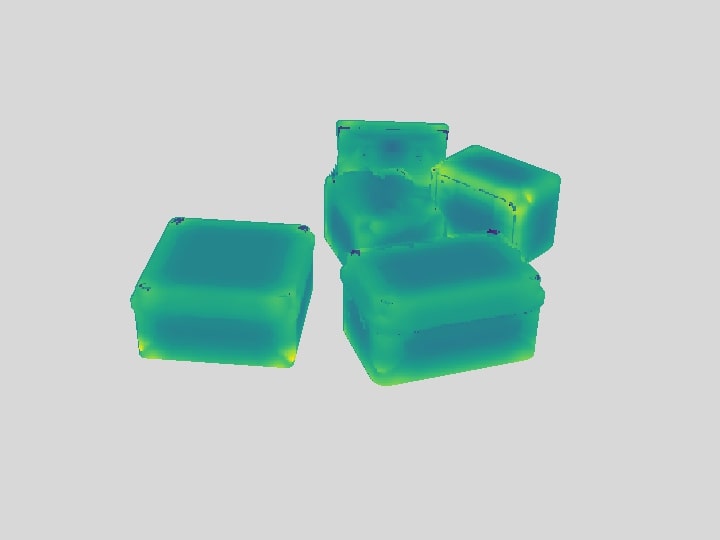} &
     \includegraphics[width=\w]{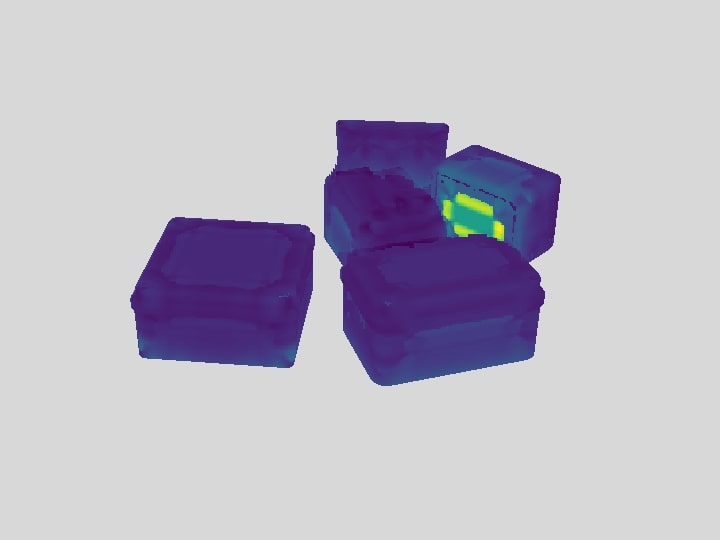} &
     \includegraphics[width=\w]{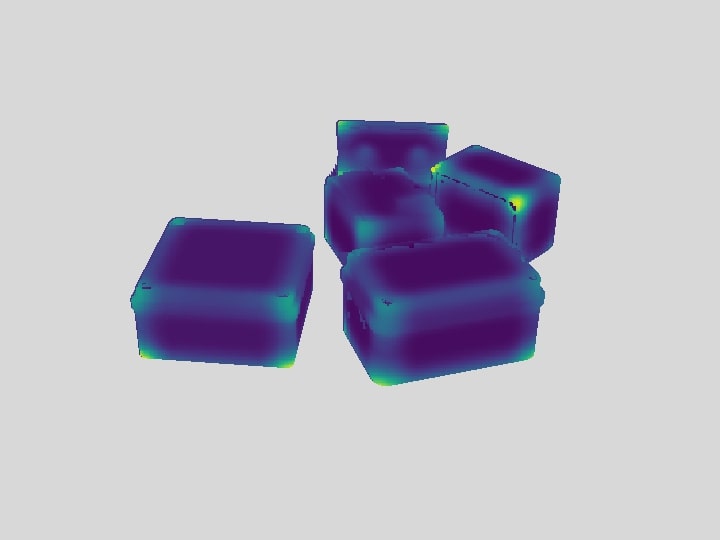} &
      \includegraphics[width=\w]{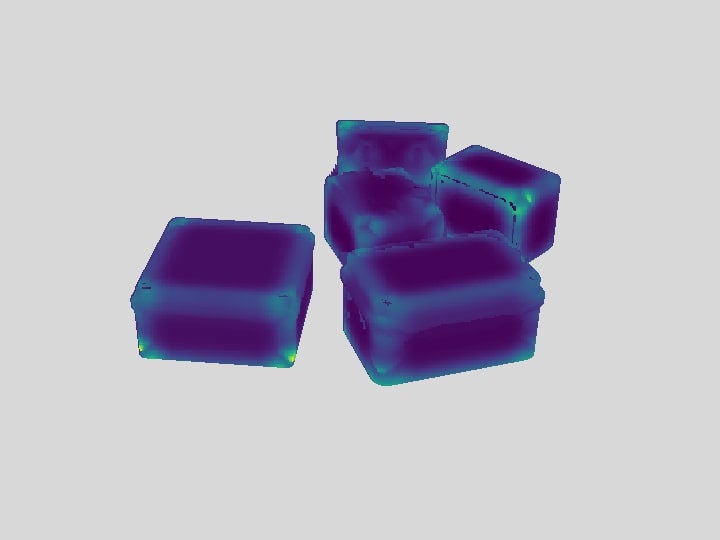} \\
     $i=2$, $j=0$, $k=1$ & $i=2$, $j=0$, $k=0$ & $i=0$, $j=0$, $k=2$ & $i=2$, $j=0$, $k=2$\\
  \end{tabular}
  \end{center}
  \caption{Visualization of some LSEs coordinates for an example image.}
\label{fig:embeddings_all_dims}
\end{figure*}

\newcommand{\imsize}{0.32\linewidth}
\begin{figure*}[t]
  \begin{center}
    \begin{tabular}{ccc}
      \includegraphics[width=\imsize]{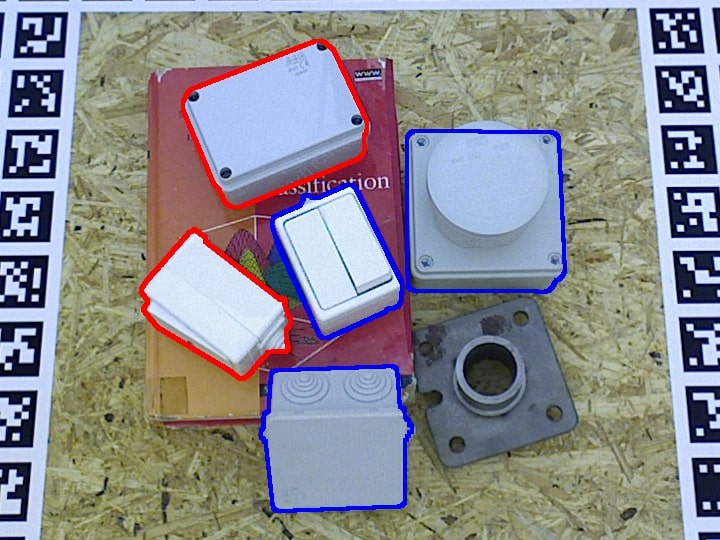} &
      \includegraphics[width=\imsize]{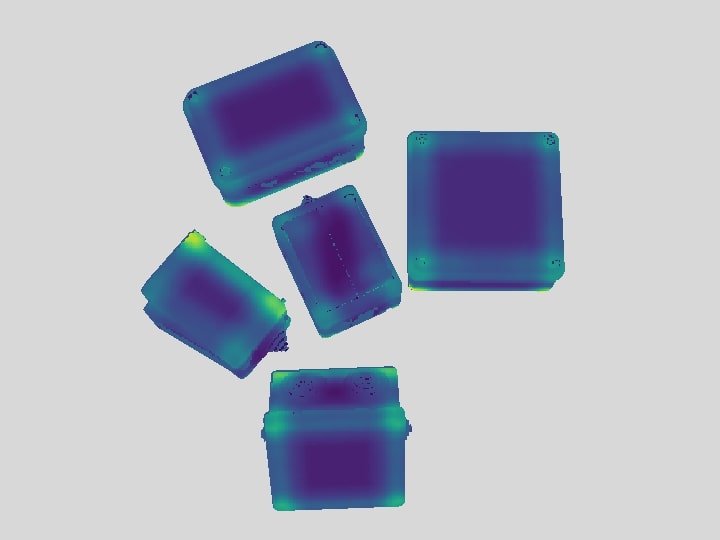} & 
     \includegraphics[width=\imsize]{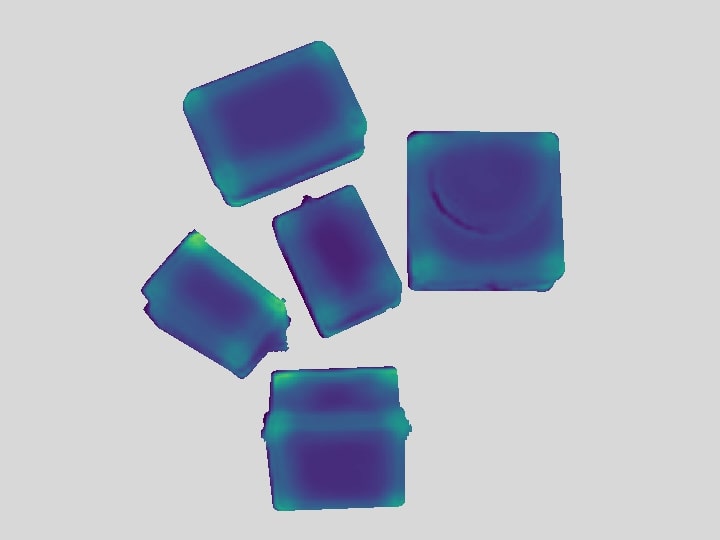} \\
      \includegraphics[width=\imsize]{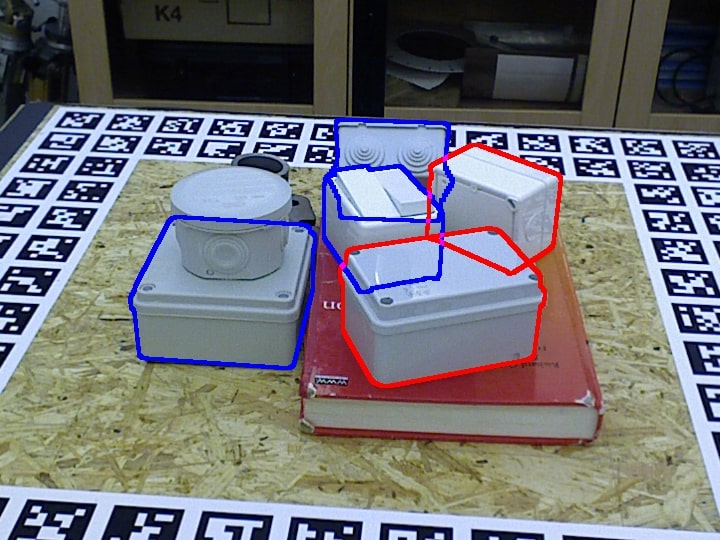} &
     \includegraphics[width=\imsize]{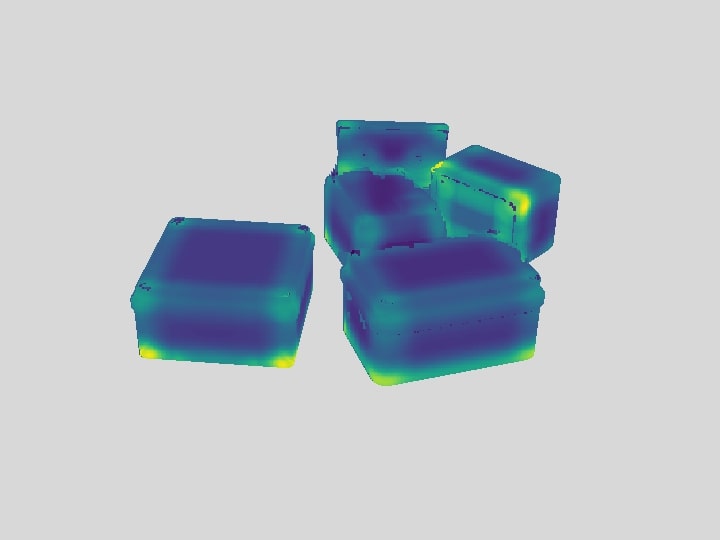} &
      \includegraphics[width=\imsize]{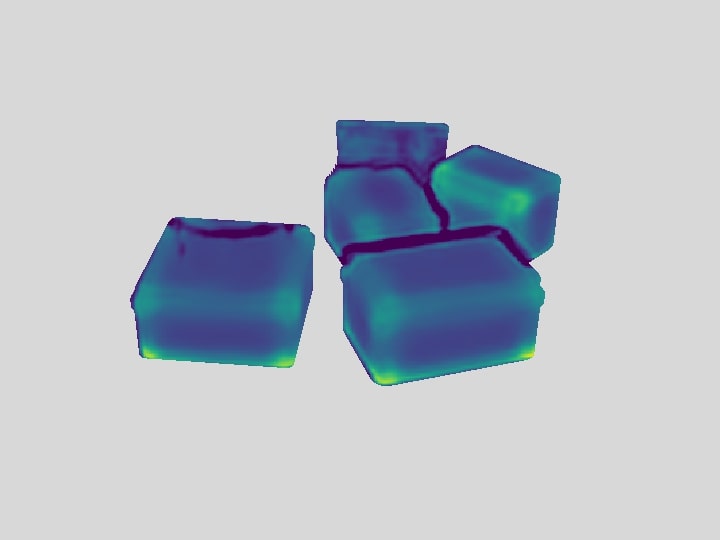} \\
      (a) & (b) & (c)\\
    \end{tabular}
  \end{center}
  \caption{Generalization of the LSE prediction network to new objects. (a) Input RGB image with objects seen during the training of the network~(blue boundaries) and new objects~(red boundaries). The LSE predictions (c) are close to the LSE Ground truth (b) for both the known and new objects. }
\label{fig:embeddings_prediction}
\end{figure*}
 
\newcommand{\calE}{\mathcal{E}}
\newcommand{\calO}{\mathcal{O}}
\newcommand{\bpp}{{\bf p}}
\newcommand{\best}{\text{best}}
\newcommand{\mini}{\text{min}}
\newcommand{\model}{C} 
\newcommand{\Models}{\mathcal{C}} 

\begin{algorithm}[t]
  \begin{algorithmic}[1]
    \State $\Models \gets$ CAD models for the new objects
    \State $\calE(\model) \gets \emb_\text{CAD}(\model)$, the LSEs of 3D points for each CAD model $\model$
    \State $I \gets$ input image
    \State $\calF \gets \emb_\text{pred}(I)$, the predicted LSEs for the input image
    \State $\calO \gets \text{Mask-RCNN}(I)$, the masks predicted by Mask-RCNN
    \State $\calM \gets \{m_i\}_i$, the set of 2D-3D matches based on $\calE(\model)$ and $\calF$. Each match $m_i$ is made of an image location $\bpp$ and 3D points on the CAD models: $(\bpp, [\bP_1, \bP_2, ..., \bP_{m_i}])$
    \State 
    \Procedure{Pose\_Estimation\_O\_C}{$O$, $\model$}
      \State $s_\best \gets 0$ 
      \For{$\text{iter} \in [0;N_\text{iter}]$}
        \State $n \gets$ random integer in $[6;10]$
        \State $M \gets  n$ random correspondences $(\bpp, \bP)$,
        \State $\quad\quad$ where $\bp\in O$ and $\bP$ is matched to $\bpp$ in $\calM$
        \State $\pose \gets \textsc{PnP}(M)$
        \State $s \gets \textsc{Score}(\pose, \model, \calE(\model), \calF, O)$
        \If{$s > s_\best$}
          \State $\pose_\best \gets \pose$
          \State $s_\best \gets s$
        \EndIf
      \EndFor 
      \State Refine $\pose_\best$
    \State \textbf{return}  $\pose_\best$, $\textsc{Score}(\pose_\best, \calE(\model), \calF, O, \model)$
    \EndProcedure
    \State 
    \Procedure{Pose\_Estimation}{}
      \For{each mask $O \in \calO$}
        \State $\vartriangleright$ $s_\mini$ is the minimum score for a match with a CAD model:
        \State $s_\best(O) \gets s_\mini$ 
        \For{each CAD model $\model$}
           \State $\pose, s \gets$ \textsc{Pose\_Estimation\_O\_C}({$O$, $C$})
           \If{$s > s_\best$}
             \State $s_\best \gets s$
             \State $\pose_\best(O) \gets \pose$
             \State $\model_\best(O) \gets C$
           \EndIf
         \EndFor
      \EndFor
    \EndProcedure
  \end{algorithmic}
  \caption{Pose estimation algorithm. }
  \label{alg:pose_estimation_algo}
\end{algorithm}

\subsection{Pose Estimation Algorithm}

The  pseudocode for our detection and pose estimation algorithm  is given  as Algorithm~\ref{alg:pose_estimation_algo}.  Given a new image, we compute the LSEs for each of its pixels using the network described in Section~\ref{sec:network} and  establish correspondences between image pixels and object 3D points. However, the number of possible correspondences can quickly become very large, which would yield a combinatorial explosion  in the number of set of correspondences needed in RANSAC. We control the complexity in two different ways.

\begin{figure}[t]
  \begin{center} 
    \begin{tabular}{cc}     \includegraphics[width=0.32\linewidth]{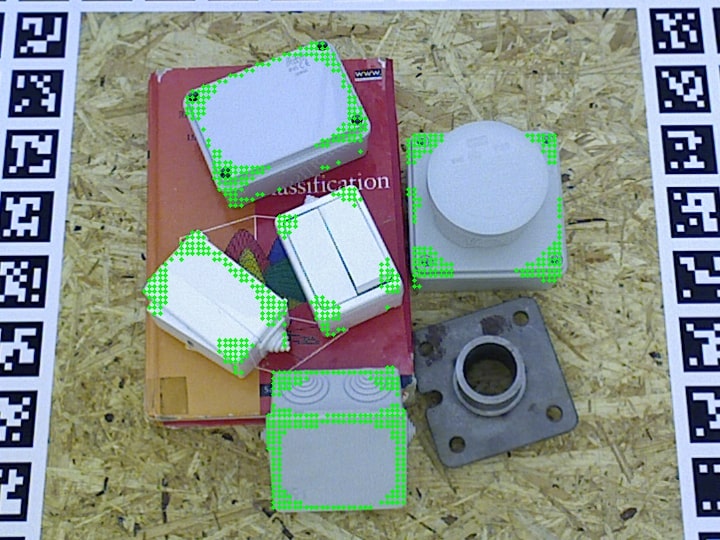} &  
    \includegraphics[width=0.57\linewidth]{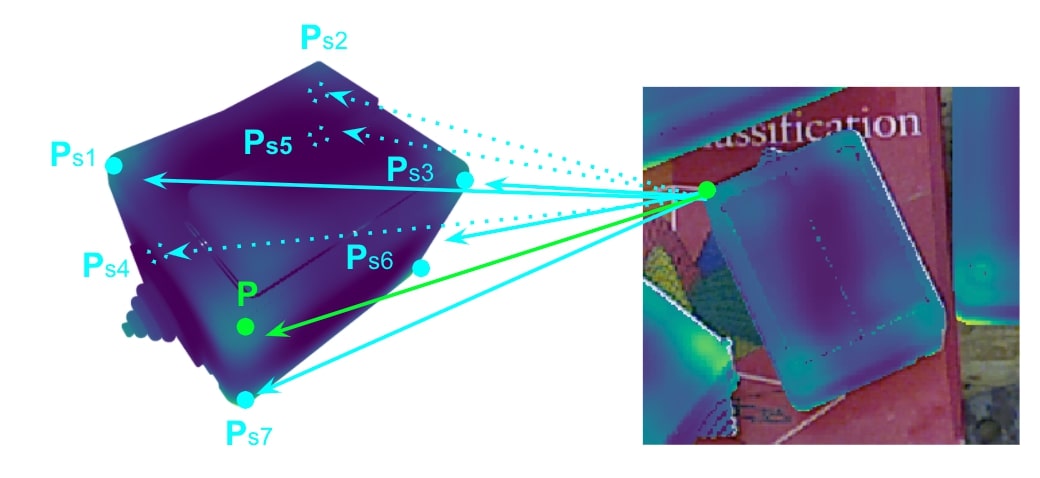}\\
    (a) & (b)
    \end{tabular}
  \end{center}
  \caption{Focusing on the most discriminative pixels. (a): In green, pixels with discriminative LSEs. We only consider them for correspondences with the CAD models. (b): A pixel can be matched with multiple 3D points on symmetrical objects because of the rotation invariance property of the LSEs.}
  \label{fig:discarding_pixels}
\end{figure}

First, we focus on the most discriminative embeddings. Points on planar regions are very common and would generate many correspondences. We discard them by thresholding the embedding values: Points with very low absolute embedding values for the LSEs are removed. Figure~\ref{fig:discarding_pixels} (a) shows how pixels are selected.


Second, we force the correspondences in each sample considered by RANSAC to belong to the same object. Even when objects are not known in advance, it is possible to segment them: To do so, we use Mask-RCNN~\cite{He17} to predict the masks of the objects. We fine-tuned it on our synthetic images already used for training the LSE predictor, as described in Section~\ref{sec:network} in a class-agnostic way since we want to generalize to new objects. We found out that it works very well with new objects even for cluttered backgrounds, as shown in Figure~\ref{fig:masks}. This also allows us to easily discard pixels on the background from the possible correspondences.


\newcommand{\littlemimsize}{0.32\linewidth}
\begin{figure*}[t]
  \begin{center}
    \begin{tabular}{ccc}
      \includegraphics[width=\littlemimsize]{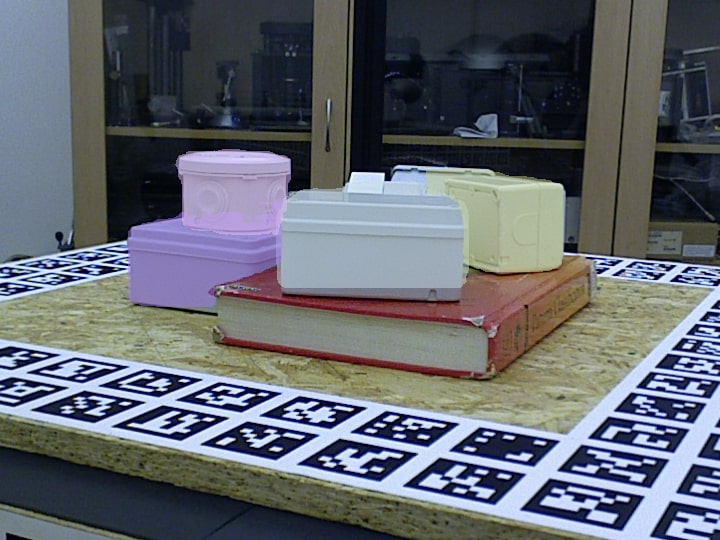} &
     \includegraphics[width=\littlemimsize]{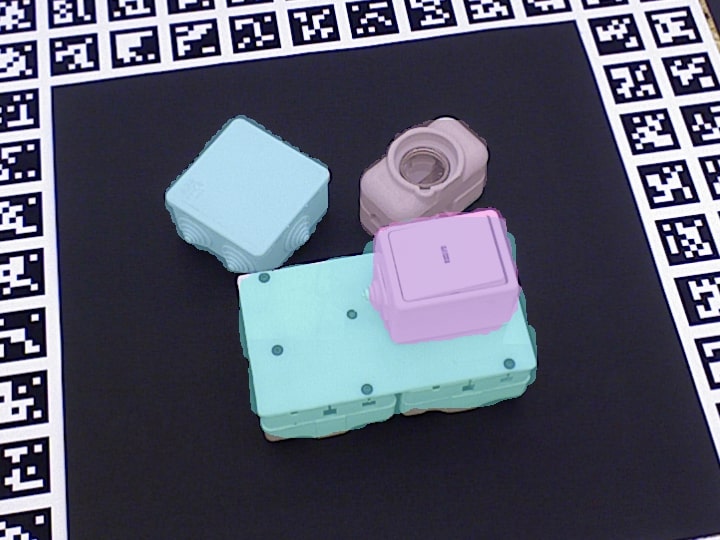} &
    \includegraphics[width=\littlemimsize]{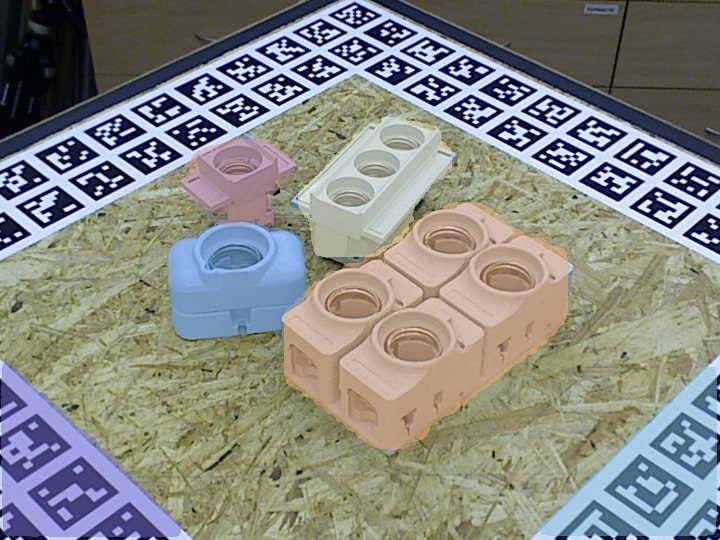} \\
    \end{tabular}
  \end{center}
\caption{Generalization of Mask-RCNN to unknown objects. We train Mask-RCNN in a class-agnostic way on a set of known objects. It generalizes well to new objects, and we use these masks to constrain the pose estimation. Note that we use masks of different colors for visualization only. Mask-RCNN cannot identify the new objects individually as it was not trained on them, it can  only detect objects in a class-agnostic way.} 
\label{fig:masks}
\end{figure*}


We match the embeddings predicted for the pixels of the input image against the  embeddings computed for the 3D points on the CAD model based on their Euclidean distances.  In our implementation, we use the FLANN library~\cite{Muja09} to efficiently get the $k$ nearest neighbors of a query embedding. In practice, we use $k=100$. This usually returns points in several clusters, as close points tend to have similar embeddings. We therefore go through the list of nearest neighbors sorted by increasing distances. We keep the first 3D point and  remove from the list the other points that are also close to this point, and we iterate. This provides for each pixel a list of potential corresponding 3D points separated from each other. 

When working on industrial objects like the ones in T-LESS, some pixels can be matched with several 3D points, as shown in Figure~\ref{fig:discarding_pixels}(b), because of the rotation invariance property of the local LSEs and the similarities between local parts of different objects. 


We finally use LO-RANSAC~(Locally Optimized RANSAC)~\cite{Chum03} with a P$n$P algorithm (we use \cite{Lepetit09} followed by a Levenberg-Marquardt optimization) to compute the poses of the visible objects. We take random $n\in[6;10]$ for each RANSAC sample. At each iteration, we compute a score for the predicted pose  as a weighted sum of  the Intersection-over-Union between the mask from Mask-RCNN and the mask obtained by rendering the model under the estimated pose, and the Euclidean distances between the predicted LSEs and the LSEs for the CAD model after reprojection. We keep the pose with the largest score and refine it using all the inlier correspondences to obtain the final 6D pose.

\section{Evaluation}
\label{sec:evaluation}

In  this section,  we present  and discuss  the results  of our  pose estimation algorithm on the challenging T-LESS dataset~\cite{Hodan17}, made of texture-less, ambiguous and symmetrical objects with no disciminative parts. It is well representative of the problems encountered in industrial context.


\subsection{Dataset}
To train our LSE prediction network, we generate synthetic images using the CAD models provided with T-LESS for a subset of objects in this dataset. The exact subset depends on the experiment, and we will detail them below. Similar to the \textit{BlenderProc4BOP} \cite{denninger2019blenderproc} introduced in the BOP challenge~\cite{hodan2018bop}, these images are created  with Cycles, a photorealistic rendering engine of the open source software Blender by randomly placing the training objects in a simple scene made of a plane randomly textured and randomly lighted. We used both these synthetics and real images for training the network combined with data augmentation to take care of the domain gap between our synthetic images and the real test images. More specifically, we use $15K$ synthetic images and $\sim7K$ real images---all the training images provided by T-LESS for the objects that are used for training the LSE prediction and Mask-RCNN. To create the ground truth embeddings, for each training image, we backproject the  pixels lying on the objects to obtain their corresponding 3D points and their LSEs. Neither the embedding prediction network nor Mask-RCNN see the test objects during training.

\subsection{LSE prediction network architecture and training} 

The architecture of the network predicting the LSEs for a given input image is a standard U-Net-like~\cite{Ronneberger15} encoder-decoder convolutional neural network taking a $720\times540$ RGB image as input. The encoder part is a 12-layer ResNet-like~\cite{He16} architecture; the decoder  upsample the feature maps up to the original size using bilinear interpolations followed by convolutional layers. We train the network with the Adam optimizer and a learning rate set to $10^{-4}$. We also use batch normalization to ensure good convergence of the model. Finally, the batch size is set to 8 and we train the network for 150 epochs.

\subsection{Metrics}
We evaluate our method using several metrics from the literature. Analogously to other related papers~\cite{zakharov2019dpod,Rad17,Pitteri19,Tekin2018}, we consider the percentage of  correctly predicted poses for each sequence  and each object of  interest, where a pose  is considered correct based on  the ADD metric~(or the ADI metric for symmetrical objects)~\cite{Hinterstoisser12}.  This  metric is based  on the average distance  in 3D between the model points after applying  the ground truth pose and the estimated one. A  pose is  considered correct  if the distance  is less  than 10\%  of the object's diameter. 

Following the BOP benchmark~\cite{hodan2018bop},  we also report the \textit{Visible Surface Discrepancy}~(VSD) metric. The VSD metric evaluates the pose error in a way that is invariant to the pose ambiguities due to object symmetries. It is computed from the distance between the estimated and ground truth visible object surfaces in the following way:
\begin{equation}
    err_\text{VSD}(\hat{S}, \bar{S}, S_I, \hat{V}, \bar{V}, \tau) =\underset{p \in \hat{V} \bigcup \bar{V}}{\text{Mean}}  \begin{cases} 
0, &\mbox{if p} \in \hat{V} \bigcap \bar{V}  \wedge |\hat{S}(p) -  \bar{S}(p)| < \tau\\ 
1, &\mbox{otherwise} 
\end{cases}
\end{equation}
where $\hat{S}$ and $\bar{S}$ are distance maps obtained by rendering the object model in the estimated and ground-truth poses respectively. The distance maps are compared with the distance map $S_I$ of the test image $I$ to obtain the visibility masks $\hat{V}$ and $\bar{V}$, {\it i.e.} the sets of pixels where the object model is visible in image $I$.
We report the mean VSD recall of 6D object poses at $err_\text{VSD} < 0.3$ with tolerance $\tau = 20mm$ and $> 10\%$ object visibility.

\subsection{Results}

The complexity  of the  test scenes in T-LESS varies from several  isolated objects  on a clean background  to very  challenging ones with  multiple instances  of several objects with a high amount of  occlusions and clutters.  We compare our method against the two works that already consider 6D object detection and pose estimation for unknown objects on T-LESS, CorNet~\cite{Pitteri19} and the MP-Encoder method of \cite{Sundermeyer20}. As the codes for these two methods are not available at the time of writing, we use the same protocols as in these works and report the results from the papers.

\begin{figure*}[t]
  \begin{center}
    \begin{tabular}{ccc}
      \includegraphics[width=\imsize]{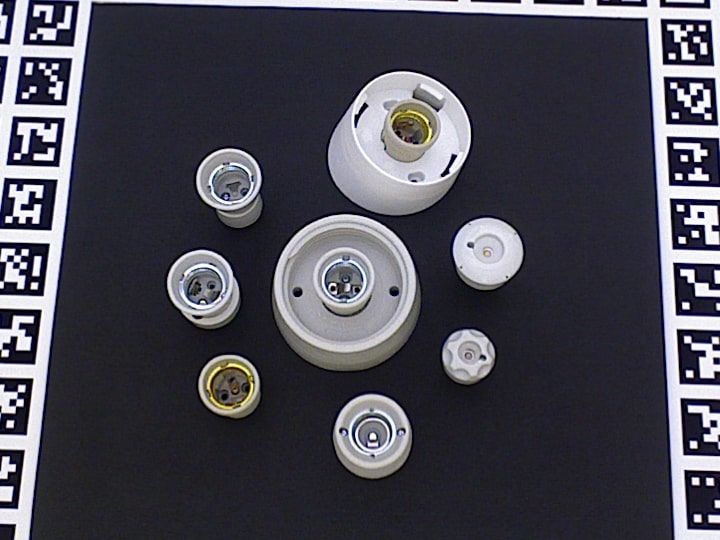} &  
     \includegraphics[width=\imsize]{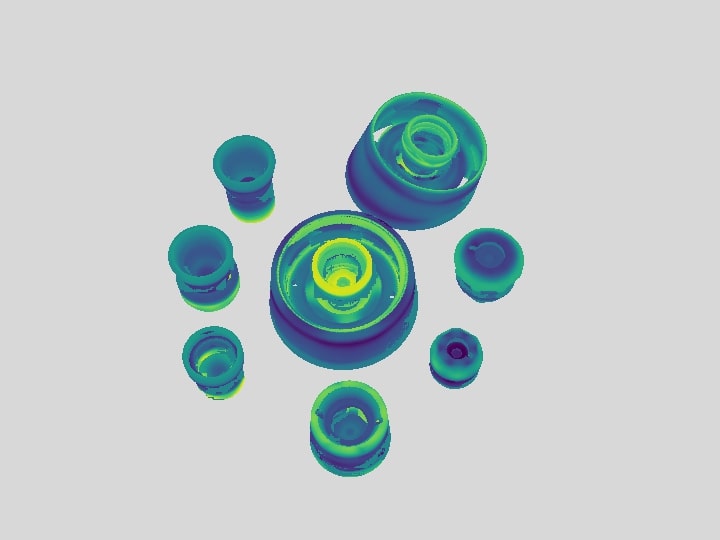} &
     \includegraphics[width=\imsize]{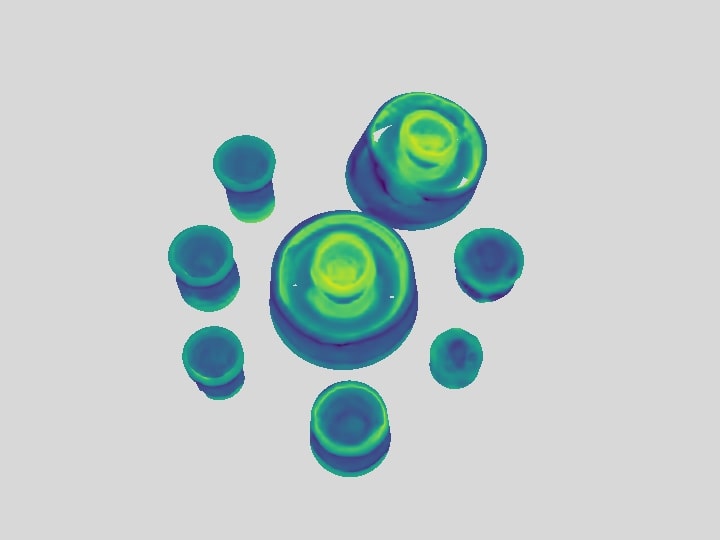} \\
     (a) & (b) & (c) \\ 
      \includegraphics[width=0.3\linewidth]{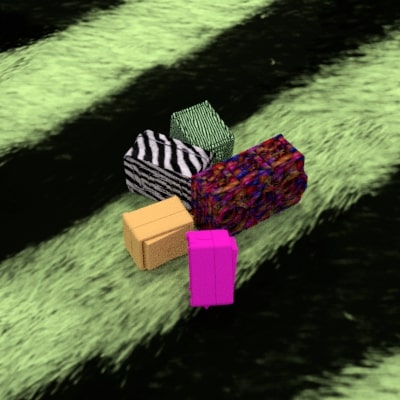} &
     \includegraphics[width=0.3\linewidth]{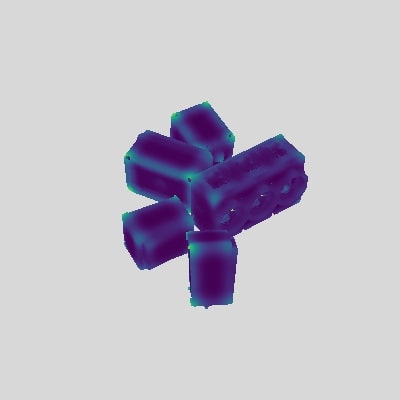} &
     \includegraphics[width=0.3\linewidth]{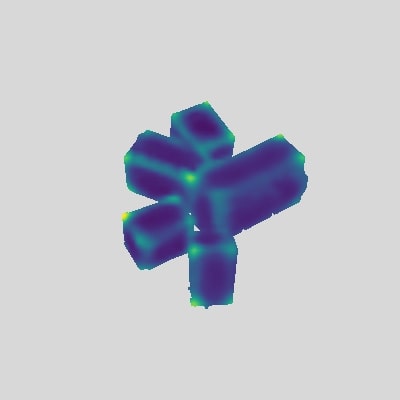} \\ 
     (d) & (e) & (f) \\
    \end{tabular}
  \end{center}
  \caption{Top row: Image with objects with rounded shapes (a). Ground truth and predicted LSEs (b) and (c). Bottom row: Image with random textures applied to some T-LESS objects (d). Ground truth and predicted LSEs (e) and (f).}
\label{fig:experiments_rounded_shape}
\end{figure*}

\paragraph{Comparison with CorNet~\cite{Pitteri19}. }
We use here the same protocol as in \cite{Pitteri19}: We split the objects from T-LESS into two sets: One set of known objects (\#6, \#19, \#25, \#27, and \#28) and one set of unknown objects (\#7, \#8, \#20, \#26, and \#29), and we compare the 3D detection and pose estimation performance of our method  and CorNet for the unknown objects in T-LESS test scenes  \#02, \#03, \#04, \#06, \#08, \#10, \#11, \#12, \#13, \#14, and \#15. We use synthetic images of the known objects for training the LSE prediction network. Results are reported in Table~\ref{tab:cornet}. We outperform CorNet on most of the objects, except on objects \#7 and \#8. This is because these objects have some 3D points with local geometry very different from the training objects (at the connections of the different parts). As a result, the predicted LSEs for these parts are not very accurate, generating wrong matches. Figured~\ref{fig:qualitative_results} shows some qualitative results for the unknown objects in the test images. 
\begin{table}[t]
\small
  \center
  \begin{adjustbox}{width=\linewidth,center}
      \begin{tabular}{l|cccccccccccccccc}
     \hline
     Scene & 02 & 03 & 04 & 04 & 06 & 08 & 10 & 11 & 11 & 12 & 12 & 13 & 15 & 15 & 14 & \\
      Obj & 7 & 8 & 26 & 8 & 7 & 20 & 20 & 8 & 9 & 7 & 9 & 20 & 29 & 26 & 20 & Avg \\
      \hline 
     \cite{Pitteri19} & \textbf{68.3} & \textbf{57.9} & 
     28.1 & 21.2 & 36.8 & 10.0 & 27.8 & \textbf{58.8} &
     - & 23.1 & - & 26.6 & 48.0 & - & 10.0 & 34.7($\pm$18.5) \\
     \textbf{Ours} & 61.0 & 44.1 & \textbf{55.6} & \textbf{39.1} & \textbf{44.8} & \textbf{38.2} & \textbf{38.3} & 40.8 & \textbf{46.1} & \textbf{41.2} & \textbf{45.8} & \textbf{39.5} & \textbf{77.0} & \textbf{63.6} & \textbf{24.9} & \textbf{46.7} ($\pm$12.0) \\
     \hline  
    \end{tabular}
    \end{adjustbox}
    \caption{\label{tab:cornet} Our quantitative results on T-LESS test Scenes \#02, \#03, \#04, \#06, \#08, \#10, \#11, \#12, \#13, \#14, \#15 as used in \cite{Pitteri19}.  We report results also  for Objects \#9 in Scenes \#11 and \#12 and for Object \#26 in Scene \#15 even though \cite{Pitteri19} does not. See text for details. }
\end{table}
\paragraph{Comparison with MP-Encoder~\cite{Sundermeyer20}. }
We use here the same protocol as in \cite{Sundermeyer20}: The objects from T-LESS are split into a set of known objects~(\#1-\#18) and one set of unknown objects~(\#19-\#30), and we compare the 3D detection and pose estimation performance of our method  and MP-Encoder for the unknown objects in T-LESS test scenes following the BOP benchmark \cite{hodan2018bop}. We use synthetic images of the known objects for training the LSE prediction network. Note that we report here the number of Table 3 from the \cite{Sundermeyer20} paper as the other reported results assume that the ground truth bounding boxes, the ground truth masks, or depth information are provided. Results are reported in Table~\ref{tab:mp-encoder}. While our method performs slightly better, the performances are close and tells us that both methods are promising. The main difference is that the MP-Encoder relies on an embedding completely learnt by a network while our method incorporates some geometrical meaning that makes our approach more appealing for industrial purposes. Note that, as shown in Fig. \ref{fig:experiments_rounded_shape}, our LSE network can handle objects with rounded shape without being limited to objects with prominent corners as \cite{Pitteri19}.

\begin{table}[t]
    \center
    \begin{tabular}{cc}
     \hline
     & VSD recall \\
     \hline
     MP-Encoder \cite{Sundermeyer20} &  20.53\\ 
     \textbf{Ours} & \textbf{23.27}\\
     \hline   
    \end{tabular}
    \caption{\label{tab:mp-encoder}Mean \textit{Visible Surface Discrepancy}~(VSD)  recall  using the protocol of \cite{Sundermeyer20}. This metric evaluates the pose error in a way that is invariant to the pose ambiguities due to object symmetries. It is computed from the distance between the estimated and ground truth visible object surfaces.}
\end{table}

\newcommand{\imagesize}{0.25\linewidth}
\begin{figure*}[] 
  \begin{center}
    \begin{tabular}{cccc} 
    \includegraphics[width=\imagesize]{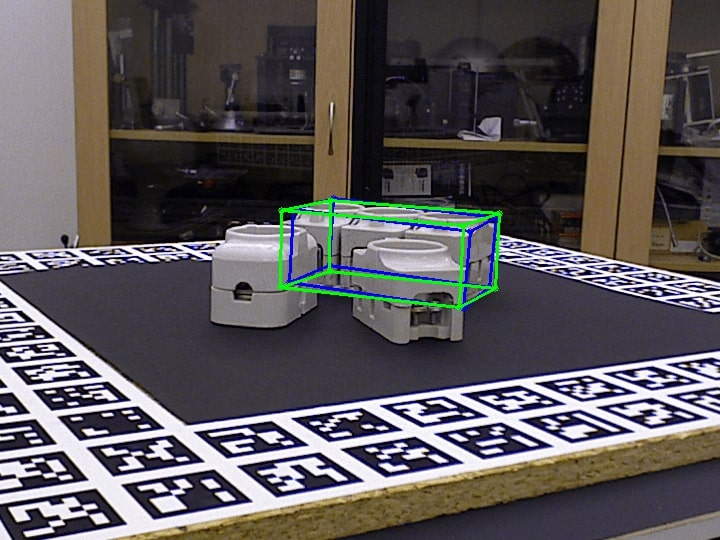} &
     \includegraphics[width=\imagesize]{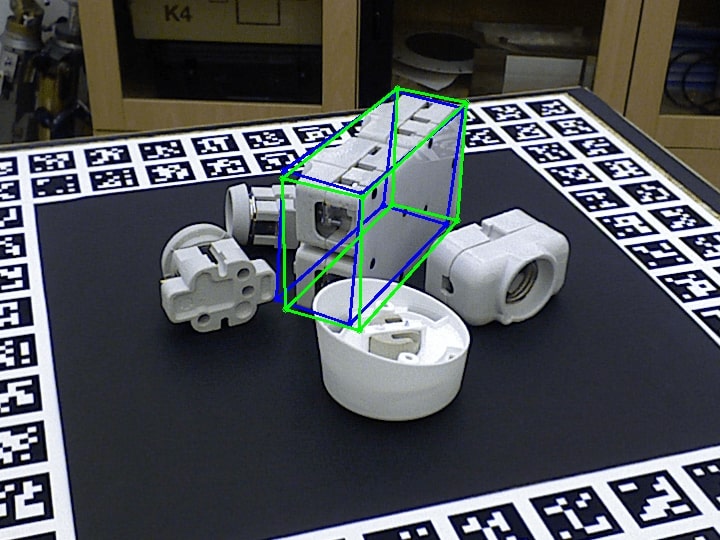} &
      \includegraphics[width=\imagesize]{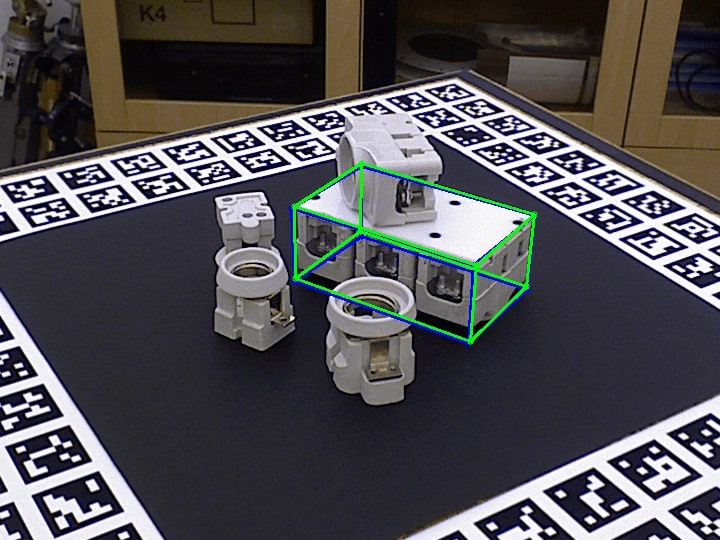} &
     \includegraphics[width=\imagesize]{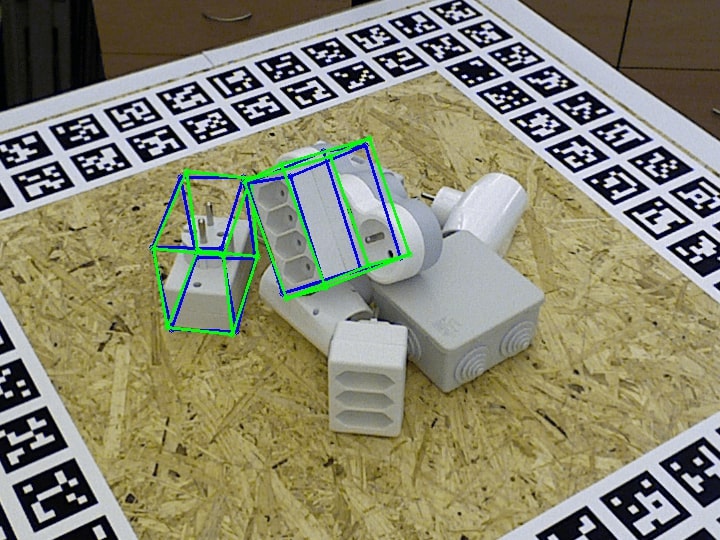} \\
     \includegraphics[width=\imagesize]{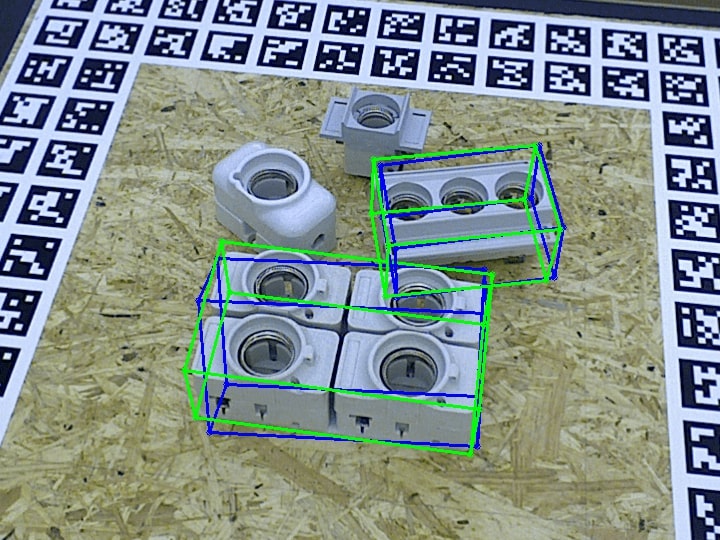} & 
     \includegraphics[width=\imagesize]{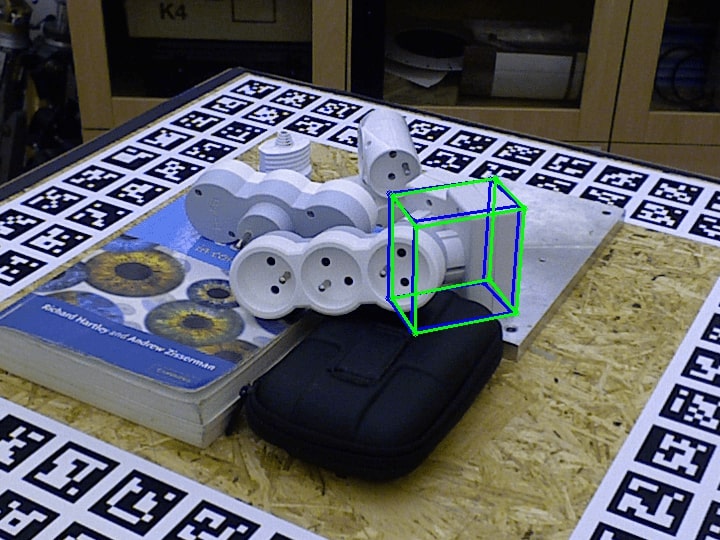} & 
     \includegraphics[width=\imagesize]{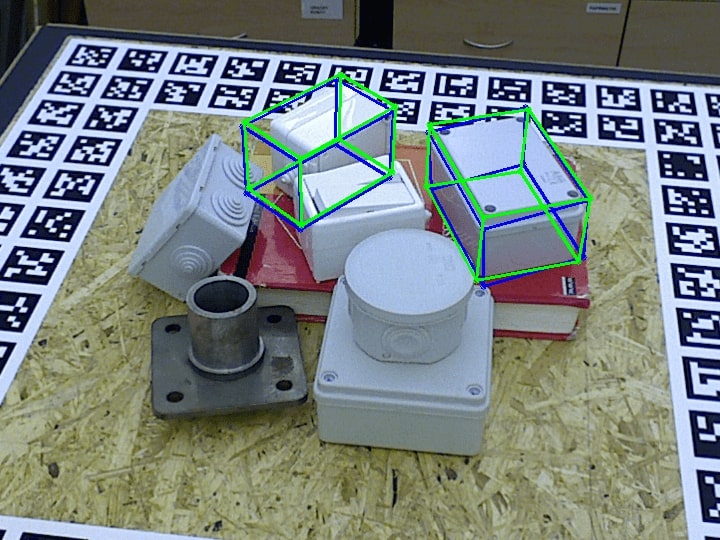} &
     \includegraphics[width=\imagesize]{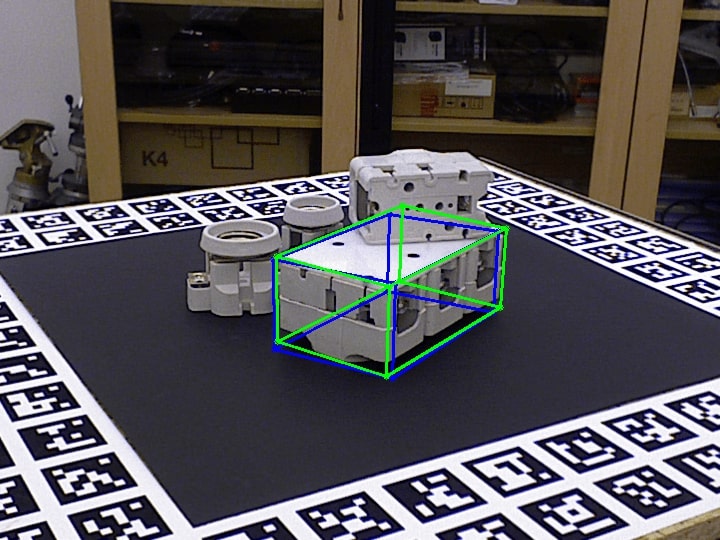} \\
     \includegraphics[width=\imagesize]{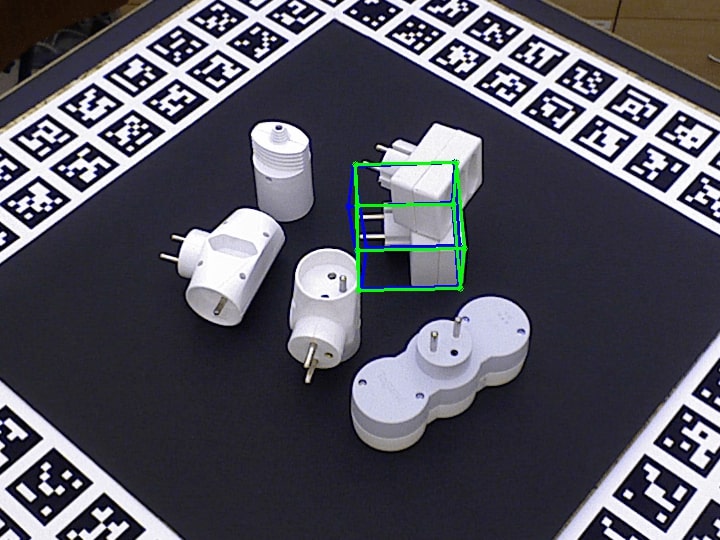} & 
     \includegraphics[width=\imagesize]{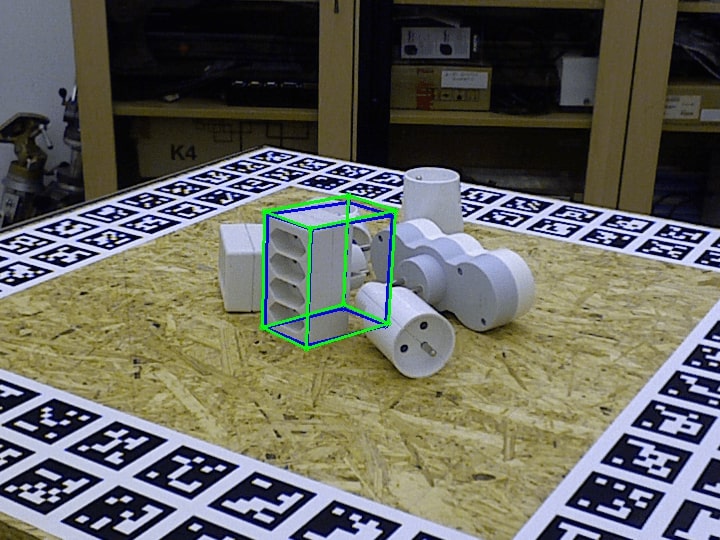} &
     \includegraphics[width=\imagesize]{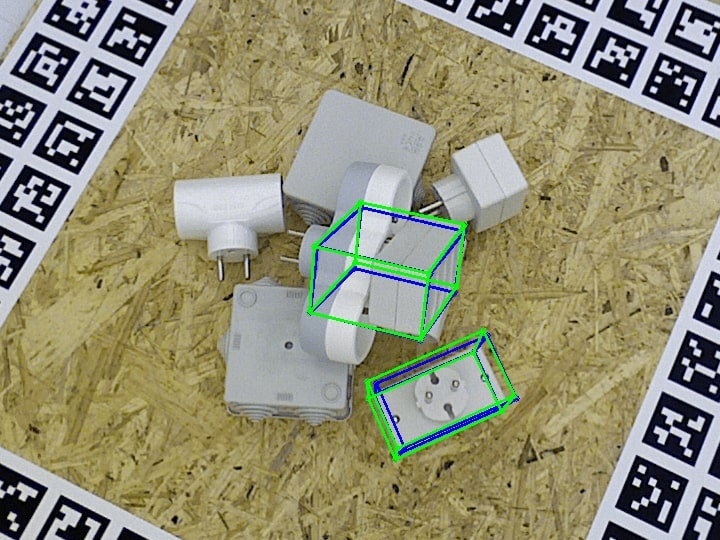}  &
     \includegraphics[width=\imagesize]{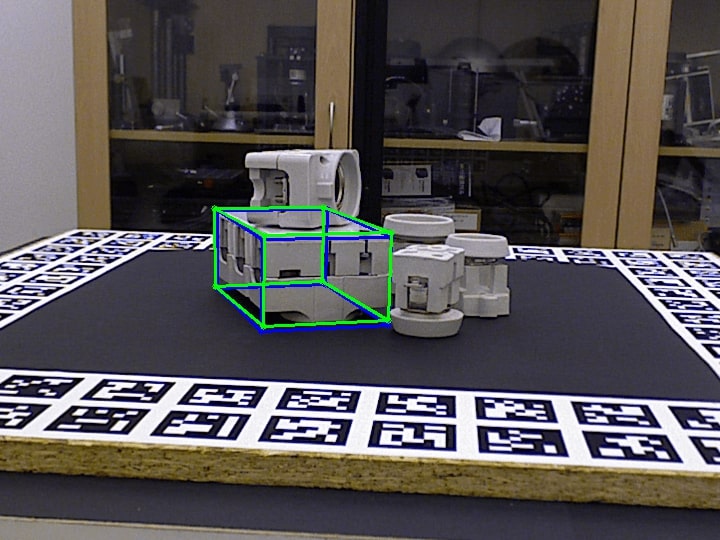}  \\
    \end{tabular}
  \end{center}
  \caption{Qualitative results on the unknown objects of the test scenes from T-LESS. Green bounding boxes denote ground truth poses, blue bounding boxes correspond to our predicted poses. } 
\label{fig:qualitative_results}
\end{figure*}

\subsection{Robustness to Texture}
Our focus is on untextured objects, as industrial objects typically do not exhibit textures like the T-LESS objects. However, our LSEs can be predicted for textured objects as well. To show this, we retrained our LSE prediction network on synthetic images of the T-LESS objects rendered with random textures. The LSEs prediction for some test images are shown in the bottom row of Fig.~\ref{fig:experiments_rounded_shape}.


\section{Conclusion}
\label{sec:conclusion}

We introduced a novel approach for the detection and the 3D pose estimation of industrial objects in color images. It only requires the CAD models of the objects and no retraining is needed for new objects. We introduce a new type of embedding capturing the local geometry of the 3D points lying on the object surface and we train a network to predict these embeddings per pixel for images of new objects. From these local surface embeddings, we establish correspondences and obtain the pose with a P$n$P+RANSAC algorithm. Describing the local geometries of the objects allows to generalize to new categories and the rotation invariance of our embeddings makes the method able to solve typical ambiguities that raise with industrial and symmetrical objects. We believe that using local and rotation invariance descriptors is the key to solve the 6D pose of new textureless objects from color images. 



\bibliographystyle{splncs}
\bibliography{string, vision, biblio, slobodan}

\begin{thebibliography}{10}

\bibitem{Kehl17}
Kehl, W., Manhardt, F., Tombari, F., Ilic, S., Navab, N.:
\newblock {SSD-6D: Making RGB-Based 3D Detection and 6D Pose Estimation Great
  Again}.
\newblock In: International Conference on Computer Vision. (2017)

\bibitem{Rad17}
Rad, M., Lepetit, V.:
\newblock {BB8: A Scalable, Accurate, Robust to Partial Occlusion Method for
  Predicting the 3D Poses of Challenging Objects Without Using Depth}.
\newblock In: International Conference on Computer Vision. (2017)

\bibitem{Tekin2018}
Tekin, B., Sinha, S.N., Fua, P.:
\newblock {Real-Time Seamless Single Shot 6D Object Pose Prediction}.
\newblock In: Conference on Computer Vision and Pattern Recognition. (2018)

\bibitem{Jafari2018}
Jafari, O.H., Mustikovela, S.K., Pertsch, K., Brachmann, E., Rother, C.:
\newblock {{IPose}: Instance-Aware 6D Pose Estimation of Partly Occluded
  Objects}.
\newblock CoRR \textbf{abs/1712.01924} (2017)

\bibitem{Xiang18}
Xiang, Y., Schmidt, T., Narayanan, V., Fox, D.:
\newblock {{PoseCNN}: A Convolutional Neural Network for 6D Object Pose
  Estimation in Cluttered Scenes}.
\newblock Robotics: Science and Systems Conference (2018)

\bibitem{zakharov2019dpod}
Zakharov, S., Shugurov, I., Ilic, S.:
\newblock {DPOD: Dense 6D Pose Object Detector and Refiner}.
\newblock In: International Conference on Computer Vision. (2019)

\bibitem{Oberweger2018}
Oberweger, M., Rad, M., Lepetit, V.:
\newblock {Making Deep Heatmaps Robust to Partial Occlusions for 3D Object Pose
  Estimation}.
\newblock In: European Conference on Computer Vision. (2018)

\bibitem{Peng18_PVNet}
Peng, S., Liu, Y., Huang, Q., Bao, H., Zhou, X.:
\newblock {PVNet: Pixel-Wise Voting Network for 6DoF Pose Estimation}.
\newblock CoRR \textbf{abs/1812.11788} (2018)

\bibitem{Hu20}
Hu, Y., Fua, P., Wang, W., Salzmann, M.:
\newblock {Single-Stage 6D Object Pose Estimation}.
\newblock In: The IEEE/CVF Conference on Computer Vision and Pattern
  Recognition. (2020)

\bibitem{sundermeyer2020augmented}
Sundermeyer, M., Marton, Z.C., Durner, M., Triebel, R.:
\newblock {Augmented Autoencoders: Implicit 3D Orientation Learning for 6D
  Object Detection}.
\newblock International Journal of Computer Vision \textbf{128} (2020)
  714--729

\bibitem{park2019pix2pose}
Park, K., Patten, T., Vincze, M.:
\newblock {Pix2pose: Pixel-Wise Coordinate Regression of Objects for 6D Pose
  Estimation}.
\newblock In: Proceedings of the IEEE International Conference on Computer
  Vision. (2019)  7668--7677

\bibitem{li2018deepim}
Li, Y., Wang, G., Ji, X., Xiang, Y., Fox, D.:
\newblock {{DeepIM}: Deep Iterative Matching for 6D Pose Estimation}.
\newblock In: European Conference on Computer Vision. (2018)  683--698

\bibitem{Pitteri19}
Pitteri, G., Lepetit, V., Ilic, S.:
\newblock {CorNet: Generic 3D Corners for 6D Pose Estimation of New Objects
  without Retraining}.
\newblock In: International Conference on Computer Vision Workshops. (2019)

\bibitem{Sundermeyer20}
Sundermeyer, M., Durner, M., Puang, E.Y., Marton, Z.C., Vaskevicius, N., Arras,
  K.O., Triebel, R.:
\newblock {Multi-path Learning for Object Pose Estimation Across Domains}.
\newblock In: Conference on Computer Vision and Pattern Recognition. (2020)

\bibitem{Brachmann16}
Brachmann, E., Michel, F., Krull, A., Yang, M.M., Gumhold, S., Rother, C.:
\newblock {Uncertainty-Driven 6D Pose Estimation of Objects and Scenes from a
  Single RGB Image}.
\newblock In: Conference on Computer Vision and Pattern Recognition. (2016)

\bibitem{NOCS_Wang2019}
Wang, H., Sridhar, S., Valentin, J.H.J., Song, S., Guibas, L.J.:
\newblock {Normalized Object Coordinate Space for Category-Level 6D Object Pose
  and Size Estimation}.
\newblock In: Conference on Computer Vision and Pattern Recognition. (2019)

\bibitem{He17}
He, K., Gkioxari, G., Dollar, P., Girshick, R.:
\newblock {Mask {R-CNN}}.
\newblock In: International Conference on Computer Vision. (2017)

\bibitem{Hodan17}
Hodan, T., Haluza, P., Obdrzalek, S., Matas, J., Lourakis, M., Zabulis, X.:
\newblock {T-LESS: An RGB-D Dataset for 6D Pose Estimation of Texture-less
  Objects}.
\newblock In: IEEE Winter Conference on Applications of Computer Vision. (2017)

\bibitem{hu2019segmentation}
Hu, Y., Hugonot, J., Fua, P., Salzmann, M.:
\newblock {Segmentation-Driven 6D Object Pose Estimation}.
\newblock In: Conference on Computer Vision and Pattern Recognition. (2019)
  3385--3394

\bibitem{peng2019pvnet}
Peng, S., Liu, Y., Huang, Q., Zhou, X., Bao, H.:
\newblock {PVNet: Pixel-Wise Voting Network for 6DoF Pose Estimation}.
\newblock In: Conference on Computer Vision and Pattern Recognition. (2019)
  4561--4570

\bibitem{Taylor12}
Taylor, J., Shotton, J., Sharp, T., Fitzgibbon, A.:
\newblock {The Vitruvian Manifold: Inferring Dense Correspondences for One-Shot
  Human Pose Estimation}.
\newblock In: Conference on Computer Vision and Pattern Recognition. (2012)
  103--110

\bibitem{Wang18}
Wang, Y., Tan, X., Yang, Y., Liu, X., Ding, E., Zhou, F., Davis, L.S.:
\newblock 3d pose estimation for fine-grained object categories.
\newblock In: European Conference on Computer Vision Workshops. (2018)

\bibitem{li2019cdpn}
:
\newblock
\newblock ({}cdpn: Coordinates-based disentangled pose network for real-time
  rgb-based 6-dof object pose estimation)

\bibitem{hodan2020epos}
Hodan, T., Barath, D., Matas, J.:
\newblock {EPOS: Estimating 6D Pose of Objects with Symmetries}.
\newblock arXiv preprint arXiv:2004.00605 (2020)

\bibitem{HinterstoisserPreTrainedImageFeatures2017a}
Hinterstoisser, S., Lepetit, V., Wohlhart, P., Konolige, K.:
\newblock {On Pre-Trained Image Features and Synthetic Images for Deep
  Learning}.
\newblock In: arXiv. (2017)

\bibitem{Goodfellow14}
Goodfellow, I.J., Pouget-Abadie, J., Mirza, M., Xu, B., Warde-Farley, D.,
  Ozair, S., Courville, A., Bengio, Y.:
\newblock {Generative Adversarial Nets}.
\newblock In: Advances in Neural Information Processing Systems. (2014)

\bibitem{Bousmalis16}
Bousmalis, K., Trigeorgis, G., Silberman, N., Krishnan, D., Erhan, D.:
\newblock {Domain Separation Networks}.
\newblock In: Advances in Neural Information Processing Systems. (2016)
  343--351

\bibitem{Mueller2018}
M\"uller, F., Bernard, F., Sotnychenko, O., Mehta, D., Sridhar, S., Casas, D.,
  Theobalt, C.:
\newblock {GANerated Hands for Real-Time 3D Hand Tracking from Monocular RGB}.
\newblock In: Conference on Computer Vision and Pattern Recognition. (2018)

\bibitem{bousmalis17}
Bousmalis, K., Silberman, N., Dohan, D., Erhan, D., Krishnan, D.:
\newblock {Unsupervised Pixel-Level Domain Adaptation with Generative
  Adversarial Networks}.
\newblock In: Conference on Computer Vision and Pattern Recognition. (2017)

\bibitem{zhu2017unpaired}
Zhu, J.Y., Park, T., Isola, P., Efros, A.A.:
\newblock {Unpaired Image-To-Image Translation Using Cycle-Consistent
  Adversarial Networks}.
\newblock In: International Conference on Computer Vision. (2017)

\bibitem{ganin2016domain}
Ganin, Y., Ustinova, E., Ajakan, H., Germain, P., Larochelle, H., Laviolette,
  F., Marchand, M., Lempitsky, V.:
\newblock {Domain-Adversarial Training of Neural Networks}.
\newblock Journal of Machine Learning Research (2016)

\bibitem{long2015learning}
Long, M., Cao, Y., Wang, J., Jordan, M.I.:
\newblock {Learning Transferable Features with Deep Adaptation Networks}.
\newblock In: International Conference on Machine Learning. (2015)

\bibitem{tzeng2015simultaneous}
Tzeng, E., Hoffman, J., Darrell, T., Saenko, K.:
\newblock {Simultaneous Deep Transfer Across Domains and Tasks}.
\newblock In: International Conference on Computer Vision. (2015)

\bibitem{lee2018diverse}
Lee, H.Y., Tseng, H.Y., Huang, J.B., Singh, M., Yang, M.H.:
\newblock {Diverse Image-To-Image Translation via Disentangled
  Representations}.
\newblock In: European Conference on Computer Vision. (2018)

\bibitem{Zakharov2018}
Zakharov, S., Planche, B., Wu, Z., Hutter, A., Kosch, H., Ilic, S.:
\newblock {Keep It Unreal: Bridging the Realism Gap for 2.5D Recognition with
  Geometry Priors Only}.
\newblock In: International Conference on 3D Vision. (2018)

\bibitem{Tobin2017}
Tobin, J., Fong, R., Ray, A., Schneider, J., Zaremba, W., Abbeel, P.:
\newblock {Domain Randomization for Transferring Deep Neural Networks from
  Simulation to the Real World}.
\newblock In: International Conference on Intelligent Robots and Systems.
  (2017)

\bibitem{Hinterstoisser12}
Hinterstoisser, S., Cagniart, C., Ilic, S., Sturm, P., Navab, N., Fua, P.,
  Lepetit, V.:
\newblock {Gradient Response Maps for Real-Time Detection of Textureless
  Objects}.
\newblock IEEE Transactions on Pattern Analysis and Machine Intelligence (2012)

\bibitem{Wohlhart15}
Wohlhart, P., Lepetit, V.:
\newblock {Learning Descriptors for Object Recognition and 3D Pose Estimation}.
\newblock In: Conference on Computer Vision and Pattern Recognition. (2015)

\bibitem{Balntas17}
Balntas, V., Doumanoglou, A., Sahin, C., Sock, J., Kouskouridas, R., Kim, T.K.:
\newblock {Pose Guided RGBD Feature Learning for 3D Object Pose Estimation}.
\newblock In: International Conference on Computer Vision. (2017)

\bibitem{zakharov2017iros}
Zakharov, S., Kehl, W., Planche, B., Hutter, A., Ilic, S.:
\newblock {3D Object Instance Recognition and Pose Estimation Using Triplet
  Loss with Dynamic Margin}.
\newblock In: International Conference on Intelligent Robots and Systems.
  (2017)

\bibitem{bui2018icra}
Bui, M., Zakharov, S., Albarqouni, S., Ilic, S., Navab, N.:
\newblock {When Regression Meets Manifold Learning for Object Recognition and
  Pose Estimation}.
\newblock In: International Conference on Robotics and Automation. (2018)

\bibitem{Eggert97}
Eggert, D., Lorusso, A., Fisher, R.:
\newblock {Estimating 3D Rigid Body Transformations: A Comparison of Four Major
  Algorithms}.
\newblock Machine Vision and Applications \textbf{9} (1997)  272--290

\bibitem{Deng18}
Deng, H., Birdal, T., Slobodan, I.:
\newblock {PPF-FoldNet: Unsupervised Learning of Rotation Invariant 3D Local
  Descriptors}.
\newblock In: European Conference on Computer Vision. (2018)

\bibitem{Ronneberger15}
Ronneberger, O., Fischer, P., Brox, T.:
\newblock {{U-Net}: Convolutional Networks for Biomedical Image Segmentation}.
\newblock In: Conference on Medical Image Computing and Computer Assisted
  Intervention. (2015)

\bibitem{Muja09}
Muja, M., Lowe, D.:
\newblock {Fast Approximate Nearest Neighbors with Automatic Algorithm
  Configuration}.
\newblock In: International Conference on Computer Vision. (2009)

\bibitem{Chum03}
Chum, O., Matas, J., Kittler, J.:
\newblock {Locally Optimized RANSAC}.
\newblock In: German Conference on Pattern Recognition. (2003)

\bibitem{Lepetit09}
Lepetit, V., Moreno-noguer, F., Fua, P.:
\newblock {{EP$n$P}: An Accurate $o(n)$ Solution to the {P$n$P} Problem}.
\newblock International Journal of Computer Vision (2009)

\bibitem{denninger2019blenderproc}
Denninger, M., Sundermeyer, M., Winkelbauer, D., Zidan, Y., Olefir, D.,
  Elbadrawy, M., Lodhi, A., Katam, H.:
\newblock Blenderproc.
\newblock arXiv preprint arXiv:1911.01911 (2019)

\bibitem{hodan2018bop}
Hodan, T., Michel, F., Brachmann, E., Kehl, W., GlentBuch, A., Kraft, D.,
  Drost, B., Vidal, J., Ihrke, S., Zabulis, X.,  et~al.:
\newblock {BOP: Benchmark for 6D Object Pose Estimation}.
\newblock In: European Conference on Computer Vision. (2018)  19--34

\bibitem{He16}
He, K., Zhang, X., Ren, S., Sun, J.:
\newblock {Deep Residual Learning for Image Recognition}.
\newblock In: Conference on Computer Vision and Pattern Recognition. (2016)

\end{thebibliography}
\end{document}